\documentclass{article}

\PassOptionsToPackage{table}{xcolor}

\usepackage[preprint]{tmlr}

\usepackage{microtype}
\usepackage{graphicx}
\usepackage{booktabs}
\usepackage{subcaption}
\usepackage{array}
\usepackage{multirow}
\usepackage{float}
\usepackage{fancyvrb}
\usepackage{amsmath}
\usepackage{amssymb}
\usepackage{mathtools}
\usepackage{amsthm}
\usepackage{tabularx}

\usepackage[
  colorlinks=true,
  linkcolor=blue,
  urlcolor=blue,
  citecolor=blue,
  pdftitle={SAILOR: Solver-Assisted Interactive LLM-based Optimization Recovery},
  pdfauthor={Shaghayegh Sadeghi, Stephen L. Smith, David C. Del Rey Fernandez},
  pdfsubject={Interactive recovery of numerical literals in optimization models},
  pdfkeywords={optimization modeling, large language models, interactive elicitation}
]{hyperref}
\newcolumntype{Y}{>{\centering\arraybackslash}X}
\newcolumntype{C}{>{\centering\arraybackslash}X}
\usepackage[most]{tcolorbox}

\definecolor{origbg}{RGB}{245,248,252}
\definecolor{maskbg}{RGB}{252,245,245}
\definecolor{rephbg}{RGB}{245,250,245}
\definecolor{boxborder}{RGB}{90,110,140}

\definecolor{maskred}{RGB}{180,50,50}
\definecolor{rephgreen}{RGB}{40,120,70}

\newtcolorbox{exampleblock}[2][]{
  enhanced,
  colback=#1,
  colframe=boxborder,
  boxrule=0.7pt,
  arc=2mm,
  left=2mm,
  right=2mm,
  top=1.5mm,
  bottom=1.5mm,
  fonttitle=\bfseries,
  title=#2
}
\theoremstyle{plain}

\theoremstyle{definition}

\theoremstyle{remark}

\newcommand{\sailor}{\texttt{SAILOR}}

\title{\sailor: Solver-Assisted Interactive LLM-based Optimization Recovery}

\author{%
\name Shaghayegh Sadeghi \email shaghayegh.sadeghi@uwaterloo.ca \\
\addr Future Cities Institute, University of Waterloo, ON, Canada
\AND
\name Stephen L. Smith \email stephen.smith@uwaterloo.ca \\
\addr Electrical and Computer Engineering, University of Waterloo, ON, Canada
\AND
\name David C. Del Rey Fern\'andez \email ddelreyfernandez@uwaterloo.ca \\
\addr Department of Applied Mathematics, University of Waterloo, ON, Canada
}

\begin{document}

\maketitle

\begin{abstract}
Natural-language descriptions of optimization problems may be incomplete or vague about numerical information that a solver requires, including costs, capacities, demands, bounds, and penalties. A language model can translate the description into code, but when a required value is absent it must either stop or guess. We present \sailor, a proof-of-concept system that detects such unsupported numerical choices, asks the user targeted follow-up questions, and updates the optimization model before returning a solution. Questions are prioritized using uncertainty and solver-derived estimates of how strongly each missing value affects the current model. We evaluate the pipeline on $1{,}723$ instances from seven masked benchmarks using an idealized simulator that returns ground-truth values. Exact objective-value agreement ranges from $27.0\%$ to $87.6\%$ across datasets, with $1.4$--$5.7$ questions per instance on average. These results establish feasibility under controlled branch-and-reveal feedback; they do not measure performance with human users or general structural model repair. Code is available at \url{https://github.com/sshaghayeghs/SAILOR}.
\end{abstract}

\section{Introduction}

Recent systems translate natural-language optimization descriptions into executable mathematical programs \citep{Wang2025LargeLM,yang2023large,kong2025alphaopt,xiao2025survey,xiao2026deepor}. These systems are often evaluated on descriptions that already contain every required number. Ordinary users may not know which quantities a formulation needs. A user might ask for the least expensive production plan while omitting overtime cost, available capacity, unmet-demand penalties, or a service-level bound. The omission need not be deliberate; the user may simply not realize that the missing quantity changes the mathematical problem.

When a required value is missing, a translator can fail, silently choose a default, or guess. A plausible guess may still produce executable code, but the resulting recommendation can answer a different problem from the one the user intended. We study whether a short clarification dialogue can repair this numerical underspecification. \sailor\ identifies literals not supported by the input, maintains approximate beliefs over them, and uses solver perturbations to estimate which omissions matter most. It asks the user about those values, updates the generated program through abstract-syntax-tree (AST) edits, and re-solves.

The intended workflow is not to require users to write complete mathematical specifications before asking for help. Instead, the system accepts an incomplete description and discovers consequential missing information during modeling. Our evaluation isolates this capability in a controlled setting: we mask values from fully specified benchmark instances and use an automated simulator to answer with the hidden benchmark values. Because every question type in that simulator ultimately discloses the exact value, the experiments test idealized recovery rather than noisy human elicitation.

\subsection*{Contributions}

\begin{itemize}
    \item We formulate interactive completion of optimization descriptions that omit required numerical information, a setting existing natural-language-to-solver benchmarks exclude by construction.
    \item We give a reusable procedure for building benchmarks in this setting: AST-level extraction of numerical constants, solver-perturbation ranking, and targeted masking that retains ground truth for automated scoring.
    \item We implement an end-to-end recovery pipeline that maintains approximate beliefs over unsupported literals, asks clarification questions, rewrites the generated program through AST editing, and re-solves.
    \item We evaluate on $1{,}723$ instances from seven masked benchmarks under an idealized answer simulator, with ablations and explicit threats to validity. Translation fidelity dominates the outcome: substituting a weaker translation model costs between $27.8$ and $54.8$ percentage points of exact agreement on the datasets we report, an effect far larger than any we could attribute to question prioritization at these sample sizes.
\end{itemize}
\section{Related Work}

\paragraph{LLM-based optimization modeling.}
Translating natural language into mathematical programs is an active area. Early benchmarks and extraction pipelines set up the task~\citep{ramamonjison2023nl4opt,ramamonjison-etal-2022-augmenting}, and later systems tackled harder problem classes: OptiMUS~\citep{ahmaditeshnizi2024optimus,ahmaditeshnizi2024optimus3} pairs structured prompting with solver feedback, Chain-of-Experts~\citep{xiao2024chainofexperts} splits formulation across specialized modules, LLMOPT~\citep{JiangShu2025llmopt} and AlphaOPT~\citep{kong2025alphaopt} learn to define and progressively improve formulations, DeepOR~\citep{xiao2026deepor} and ORLM~\citep{huang2025orlm} train dedicated models, and OptimAI~\citep{thind2025optimai} together with global constraint agents~\citep{cai2025global} adopt agentic pipelines. LLMs have also been applied to diagnosing infeasible models~\citep{chen2024diagnosing}, supply chain optimization~\citep{li2023large}, and mathematical discovery through program search~\citep{romera2024mathematical}; surveys cover the wider landscape~\citep{xiao2025survey,Wang2025LargeLM}. Because translation quality drives our pipeline, we also build on reasoning-oriented prompting~\citep{wei2022chain,wang2022self,chen2022program,yao2023tree,besta2024graph} and on neurosymbolic methods that ground model outputs in formal tools~\citep{yao2023react,olausson-etal-2023-linc,pan-etal-2023-logic,ye2023satlm}. These systems generally evaluate fully specified inputs; we study missing or vague numerical literals after translation.

\paragraph{Self-correction and iterative refinement.}
One alternative to asking the user is to let the LLM revise its own guesses. Self-Refine~\citep{madaan2023self} and Reflexion~\citep{Shinn2023Reflexion} report benefits from iterative feedback, while \citet{huang2023large} and \citet{valmeekam2023can} identify limits without external grounding. In \sailor, optional refinement follows a revealed benchmark value rather than free-running self-critique (Section~\ref{sec:answer-processing}).

\paragraph{Interactive elicitation and active learning.}
Choosing maximally informative queries has a long history in active learning~\citep{settles2009active} and preference elicitation~\citep{viappiani2010optimal}, with Bayesian optimization~\citep{snoek2012practical} providing a framework for sequential experimental design under uncertainty and recent work linking active query selection to preference modeling for language models~\citep{melo2024deep}. Our criterion combines expected entropy reduction with solver-derived importance, estimated by perturbing and re-solving the translated model rather than by a learned surrogate.

\paragraph{Clarification questions and decision-focused learning.}
Resolving underspecified intent by asking has been studied in conversational text-to-SQL, where multi-turn benchmarks such as SParC~\citep{yu-etal-2019-sparc} and CoSQL~\citep{yu-etal-2019-cosql} formalize settings in which systems request missing information, and in ambiguity-aware code generation~\citep{li2023python}; we apply the idea to numerical literals in optimization models. Our importance scoring further draws on decision-focused learning, which favors information that affects a downstream optimization outcome over information that only lowers prediction error~\citep{elmachtoub2022smart,wilder2019melding,donti2017task}.
\section{Problem Statement}

Consider an optimization problem submitted in natural language by a user who does not know which numerical details the formulation requires. Let \( I_0 \) be the description the system initially receives. Some values needed to instantiate the intended problem are missing, ambiguous, or only implied. These might be costs, capacities, demands, bounds, penalty coefficients, or other problem-defining constants. We assume the user can clarify a value when the system asks a concrete question, even though the user did not know to include it in the initial description.

Let \( I^\ast \) be the user's fully specified intended description and \( \boldsymbol{\theta}^{\ast} = (\theta_1^{\ast}, \dots, \theta_n^{\ast}) \) the latent vector of true hidden parameters behind it. For the formal problem we assume \( I_0 \) fixes the combinatorial and structural template and that \( \boldsymbol{\theta}^{\ast} \) supplies only missing numbers; the implemented translator does not necessarily preserve this assumption, and structural translation errors lie outside the recovery mechanism.

For a candidate parameter vector \( \boldsymbol{\theta} \), let \( \mathcal{P}(\boldsymbol{\theta}) \) be the corresponding fully specified problem, with objective \( f(x;\boldsymbol{\theta}) \), feasible region \( \mathcal{X}(\boldsymbol{\theta}) \), and optimal value \( z^{\ast}(\boldsymbol{\theta}) = \min_{x \in \mathcal{X}(\boldsymbol{\theta})} f(x;\boldsymbol{\theta}) \). Because \( \boldsymbol{\theta}^{\ast} \) is unknown, the system must infer it through interaction: each round it asks a clarification query about one or more hidden parameters and updates its estimate from the answer, emitting a recovered vector \( \tilde{\boldsymbol{\theta}}^{(K)} \) after at most \( K \) rounds.

We measure recovery by comparing objective values. In the experiments the primary target is the benchmark answer used by the aggregation pipeline; comparing instead against the translated formulation's own re-solved objective is an internal diagnostic, not an independent target. The normalized objective error is

\begin{equation}
\mathrm{ErrObj}\!\left(\tilde{\boldsymbol{\theta}}^{(K)}, \boldsymbol{\theta}^{\ast}\right)
=
\frac{\left| z^{\ast}\!\left(\tilde{\boldsymbol{\theta}}^{(K)}\right) - z^{\ast}\!\left(\boldsymbol{\theta}^{\ast}\right) \right|}
{\max\!\left( \left| z^{\ast}\!\left(\boldsymbol{\theta}^{\ast}\right) \right|, 1 \right)}.
\end{equation}

This is an objective-value agreement criterion, not strict parameter recovery and not a guarantee that the intended optimizer has been recovered: different parameter vectors can yield the same objective value, and objective coefficients can change both the value and the optimizer.

\vspace{2mm}
\noindent
\textbf{Problem 1.}
Given an underspecified description \( I_0 \), an unknown latent vector \( \boldsymbol{\theta}^{\ast} \), and a budget of \( K \) clarification rounds, find a sequence of queries and updates yielding a recovered vector \( \tilde{\boldsymbol{\theta}}^{(K)} \) that minimizes \( \mathrm{ErrObj}(\tilde{\boldsymbol{\theta}}^{(K)}, \boldsymbol{\theta}^{\ast}) \).

\section{Method}

\subsection{Overview}

\sailor\ translates an underspecified description $I_0$ into executable Gurobi code together with a record of every number it had to guess. It separates constants stated in the text from constants it inferred, places a belief over the inferred ones, and estimates how much each currently matters to the optimization. It then asks clarification questions that score well on both information gain and solver relevance; each answer triggers a belief update, a code rewrite, and a re-solve. The loop stops once the hidden parameters are resolved or the budget runs out.

\begin{figure}[htb]
    \centering
    \includegraphics[width=\linewidth]{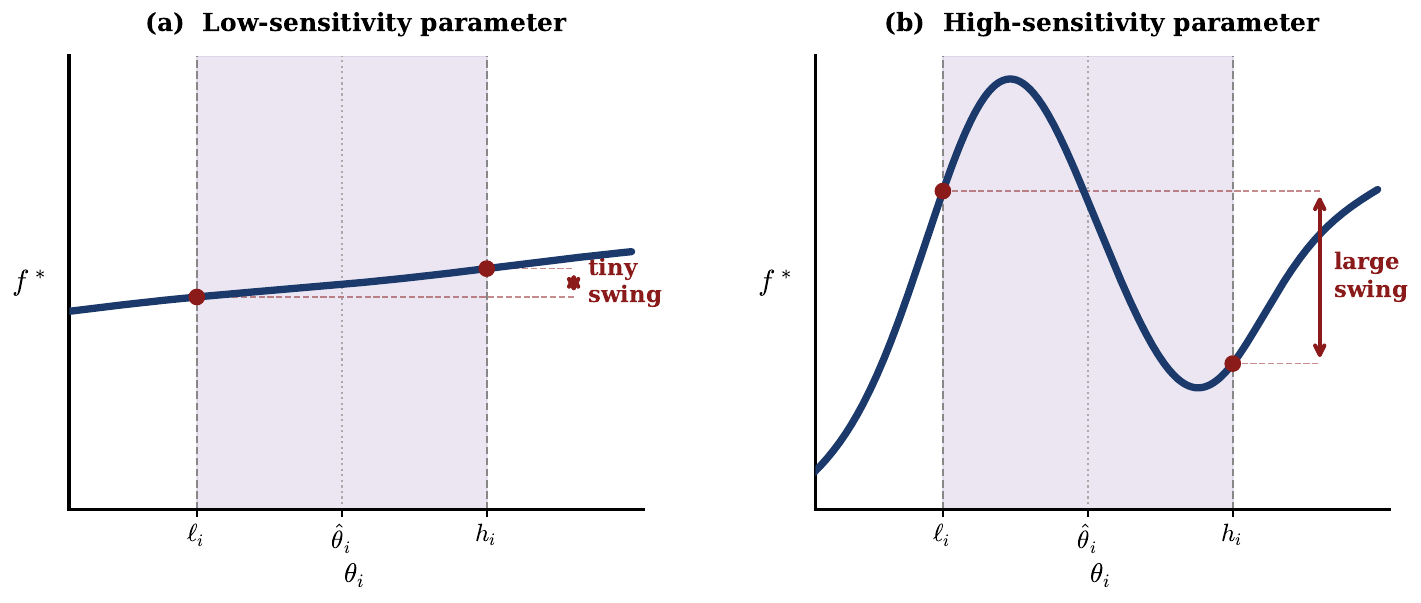}
\caption{Parameter sensitivity under uncertainty. In panel~(a), perturbing a low-sensitivity parameter across its current belief range barely moves the optimal objective. In panel~(b), the same kind of sweep over a high-sensitivity parameter swings the objective sharply. That gap is why we use solver-based sensitivity to decide which parameters to ask about first.}
\label{fig:sensitivity}
\end{figure}

\subsection{LLM Translation}

At initialization, a large language model translates the masked description $I_0$ into an executable formulation plus a record of introduced numbers and their semantic roles. Failed validation can trigger repair attempts, and a separate verification call can replace the initial code. Once initialization succeeds, interaction operates on the generated representation and extracted parameter set rather than retranslating the original text each round.
\subsection{Parameter Identification}

We parse the generated Python program with the abstract syntax tree (AST) module. Each numerical literal in the code is sorted into one of two categories:
\begin{itemize}
    \item \textbf{Known}: the value appears in $I_0$, so no clarification is needed.
    \item \textbf{Guessed}: the value is absent from $I_0$ and the LLM supplied it.
\end{itemize}
The translator returns a list of guessed parameters. AST constants are associated with that list by value and, when necessary, name similarity. Separately, numeric tokens extracted by a regular expression from $I_0$ identify literals explicitly present in the text, allowing numerically equivalent decimal forms such as ``$40$'' and ``$40.0$.'' Our matcher does not normalize number words such as ``forty,'' and ambiguous matches can remain unresolved. The output is an ordered set of extracted parameters with their values and statuses.

\subsection{Gaussian Belief Model}

Each unresolved parameter $\theta_i$ gets an approximate Gaussian belief,
\begin{equation}
\theta_i \sim \mathcal{N}(\mu_i, \sigma_i^2), \qquad \theta_i \in [\ell_i, h_i],
\end{equation}
with $\mu_i$ initialized from the LLM's guess. For the positive-valued parameters used by the default implementation the initial interval is \( [\ell_i, h_i] = [0.01\,\mu_i,\, 100\,\mu_i] \), and the initial standard deviation is proportional to its width.

Three update operations are defined. An \emph{exact update} takes a revealed value and collapses the posterior to a point mass, $\sigma_i \rightarrow 0$. A hypothetical \emph{pure binary update} would retain only the branch selected by a threshold answer and re-center the approximation. A \emph{soft update} shifts the mean toward an optional LLM refinement with a fixed blending weight.

Residual uncertainty is measured by differential entropy,
\begin{equation}
H_i = \frac{1}{2}\log\!\left(2\pi e\,\sigma_i^2\right),
\end{equation}
and resolved parameters are assigned a floor of $-50$ to stand in for negligible remaining uncertainty.

\paragraph{Design choice.} A raw-scale Gaussian gives a simple entropy surrogate but is poorly matched to several common parameter types: the multiplicative interval is meaningful only for positive, nonzero guesses, untruncated mass can violate known bounds, continuous beliefs ignore integrality, and the entropy expression is only approximate after truncation. Log-normal or truncated distributions would suit positive scales better, while signed, bounded, and integer parameters need type-specific supports. The implementation does not resolve this, so the prior-range experiment is a limited sensitivity check.

\begin{figure}[htb]
    \centering
    \includegraphics[width=\textwidth]{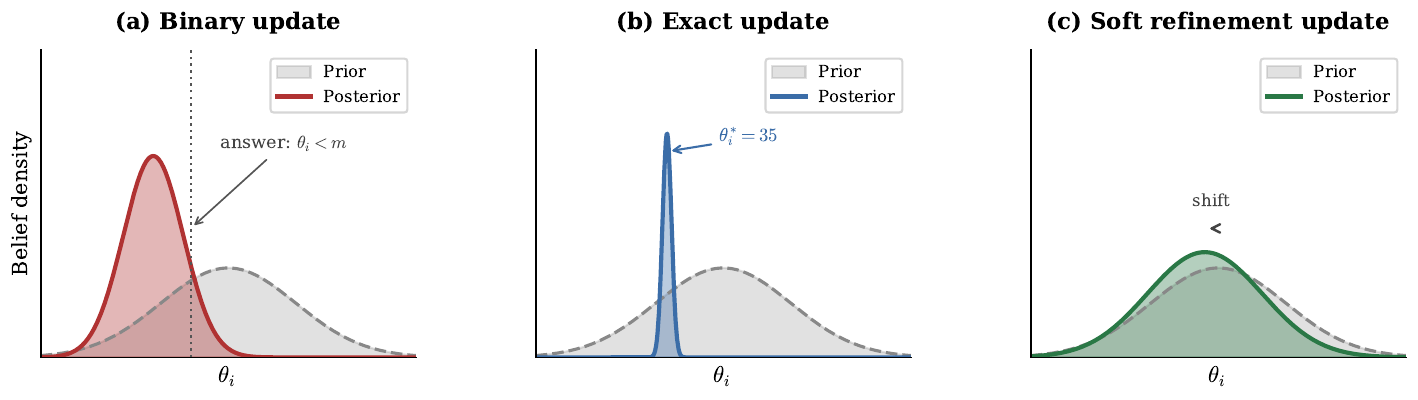}
    \caption{The three conceptual belief-update operations. Each panel shows the prior (dashed gray) and posterior (solid). Panel~(a) is a pure binary update, panel~(b) an exact update, and panel~(c) an optional soft refinement update. Our experiments exercise (b) and (c); see Section~\ref{sec:setup}.}

    \label{fig:belief_update}
\end{figure}

\subsection{Importance Scoring}

We also estimate how relevant each unresolved parameter currently is to the optimization. For each $\theta_i$, the system perturbs the value across the current belief range, re-solves, and measures the effect. The importance score is
\begin{equation}
s_i
=
w_{\mathrm{sw}} \, \phi_{\mathrm{sw}}(\theta_i)
+
w_{\mathrm{gr}} \, \phi_{\mathrm{gr}}(\theta_i)
+
w_{\mathrm{sh}} \, \phi_{\mathrm{sh}}(\theta_i)
+
w_{\mathrm{bd}} \, \phi_{\mathrm{bd}}(\theta_i),
\end{equation}
where $\phi_{\mathrm{sw}}(\theta_i)$, $\phi_{\mathrm{gr}}(\theta_i)$, $\phi_{\mathrm{sh}}(\theta_i)$, and $\phi_{\mathrm{bd}}(\theta_i)$ estimate normalized objective swing, local finite-difference sensitivity, maximum related shadow-price magnitude, and the fraction of related constraints currently binding. These are numerical diagnostics of the translated model rather than exact sensitivities of the intended problem. The weights are fixed and nonnegative, and scores are normalized to $[0,1]$ across the unresolved set when available.

We recompute importance after every interaction step. Resolving one parameter can reshape the local sensitivity structure of what remains.

\paragraph{Mixed-integer and nonlinear models.}
Shadow prices and binding-constraint sets are cleanest for linear programs. Our executor reads dual and slack attributes when the solver makes them available and otherwise sets those score components to zero. We do not construct a separate LP relaxation. Objective swings and finite differences remain available but can be unstable near infeasibility, discontinuities, or solver tolerances. None of the four components is an exact sensitivity certificate.

\subsection{Question Generation and Scoring}

Each round, the system generates candidate questions for every unresolved parameter. \emph{Direct} questions request the exact value. \emph{Binary range-split} questions ask whether the value is above or below a threshold, and \emph{confirmation} questions test a current estimate.

A candidate question $q$ for parameter $\theta_i$ is scored by its expected information gain, scaled by how relevant that parameter currently is:
\begin{equation}
\mathrm{score}(q) = \Delta H(q)\, g(s_i),
\end{equation}
where $s_i$ is the importance score of $\theta_i$ and
\begin{equation}
\Delta H(q) = H_i - \mathbb{E}\!\left[ H_i \mid \text{answer to } q \right]
\end{equation}
is the expected entropy reduction. For binary questions the expectation runs over the two posterior branches, above and below the threshold. For direct questions the score comes from the entropy reduction of resolving the parameter outright, optionally scaled by a constant to reflect the extra value of an exact answer.

We want $g(\cdot)$ monotone increasing, so that more relevant parameters get priority without uncertainty dropping out of the picture. We use the linear form
\begin{equation}
g(s_i) = \alpha + (1-\alpha)s_i,
\end{equation}
with $\alpha = 0.5$ unless stated otherwise.

The next question is
\begin{equation}
q^{(t+1)} \in \arg\max_{q \in \mathcal{Q}^{(t)}} \mathrm{score}(q).
\end{equation}

This favors questions that are informative and consequential at the same time. Note that importance only ranks questions. It never resolves, fixes, or discards a parameter on its own.

\subsection{Answer Processing and Code Rewriting}
\label{sec:answer-processing}

Each response triggers five steps. The system (i)~updates the belief over the targeted parameter, (ii)~rewrites the corresponding literal through AST editing, (iii)~re-solves the Gurobi model, (iv)~recomputes importance estimates for unresolved parameters, and (v)~optionally asks the LLM to refine its remaining guesses. Refinements enter through the soft update described above.

\subsection{Stopping Criteria}

The loop halts on any of four conditions. (i)~Every unresolved parameter has posterior standard deviation below $10^{-8}$. (ii)~Total differential entropy over unresolved parameters drops below $H_{\min}$. (iii)~Entropy fails to improve by more than $0.1$ and no parameter is filled for $N_{\text{stall}}$ consecutive rounds. (iv)~The per-instance budget $K$ is spent. We use $H_{\min}=-40$, $N_{\text{stall}}=5$, and $K=30$.

\section{Benchmark Construction}

We build benchmarks from fully specified problems whose ground truth we know but hide during recovery. The procedure runs as follows. Start with a complete instance $I^*$ where every number is stated. An LLM translates $I^*$ into executable Gurobi code. We extract the numerical constants by AST parsing plus model inspection. We then run perturbation-based sensitivity analysis, varying each parameter by $\times\{0.5, 0.8, 1.2, 2.0\}$ and re-solving. Objective sensitivity, solution sensitivity, and feasibility impact together give a structural importance ranking. We mask the top-ranked parameters with vague phrasing such as ``a certain rate'' or ``some quantity,'' and may paraphrase the result so it reads naturally.

Building benchmarks this way keeps the evaluation focused on hidden parameters that genuinely move feasibility or the optimal solution, rather than on incidental constants.

\paragraph{Caveat: shared sensitivity signals.}
The signals we use to pick parameters for masking (objective sensitivity, solution sensitivity, feasibility impact) overlap heavily with the signals \sailor's importance scorer uses at recovery time. The benchmark therefore tests \sailor\ on exactly the parameters its own scorer is built to find. We flag this circularity here and return to it, along with more independent masking criteria, in Section~\ref{sec:limitations}.

\section{Experiments}

\subsection{Experimental Setup}
\label{sec:setup}
\paragraph{Data.}
The source collection contains $1{,}732$ instances from seven established
benchmarks covering a wide difficulty range. These run from simple
two-variable LPs (NL4OPT) and mid-sized linear programs (NL4LP,
ReSocratic, EasyLP) to multi-stage industrial problems (IndustryOR,
ComplexOR) and complex LPs with up to $87$ interacting parameters
(ComplexLP), spanning both small mostly well-specified problems and the
harder regime where many decision-critical parameters must be inferred.
Table~\ref{tab:datasets} summarizes the benchmarks \citep{xiao2025survey}.
We evaluate $1{,}723$ instances end to end, excluding nine---seven from ReSocratic, one from IndustryOR, and one from ComplexLP---because translation or the oracle solve failed.

\begin{table*}[t]
\centering
\small
\renewcommand{\arraystretch}{1.15}
\setlength{\tabcolsep}{6pt}
\caption{Datasets used in our evaluation. \textit{Avg Params} is the mean
number of numerical parameters per instance, taken from the ground-truth
optimization model.}
\label{tab:datasets}
\begin{tabularx}{\linewidth}{@{}l r r l X@{}}
\toprule
\textbf{Dataset} & \textbf{Instances} & \textbf{Avg Params}
  & \textbf{Source} & \textbf{Description} \\
\midrule
NL4OPT      & 228  & 3.5         & \citet{ramamonjison2023nl4opt}
  & Simple two-variable linear programs. \\
NL4LP       & 178  & 3.9         & \citet{xiao2025survey}
  & Linear programs of moderate complexity. \\
ReSocratic  & 403  & $\sim\!3$   & \citet{xiao2025survey}
  & Socratic-style optimization problems. \\
EasyLP      & 652  & $\sim\!4$   & \citep{huang2024mamo}
  & Small linear programs from MAMO. \\
ComplexLP   & 211  & 8.4         & \citep{huang2024mamo}
  & Complex LPs, up to $87$ parameters. \\
IndustryOR  & 42   & 5.8         & \citet{xiao2025survey}
  & Industrial scheduling, $15$--$30$ parameters. \\
ComplexOR   & 18   & 2.2         & \citet{xiao2024chainofexperts}
  & Mixed-integer, nonlinear, multi-stage. \\
\midrule
\textbf{Total} & \textbf{1{,}732} & -- & -- & -- \\
\bottomrule
\end{tabularx}
\end{table*}
\paragraph{Models and infrastructure.}
We use GPT-5.2 for benchmark masking and GPT-5.3-Codex for recovery translation, verification, repair, and optional refinement. GPT-4o-mini handles answer interpretation and the simulator's extraction fallback. Requests use temperature $0$ except optional masking paraphrases, which use $0.3$. We solve with Gurobi~11 under a $30$-second per-solve limit on an Apple M-series machine with $32$ GB of RAM.

\paragraph{Evaluation protocol.}
Evaluation is fully automated through a simulated user. The simulator knows the benchmark values; the recovery system does not. Direct questions return the exact value. Binary questions return the branch and then reveal the exact value; confirmation questions similarly return a verdict and the exact value. All evaluated question types therefore resolve the targeted literal through an exact update (Appendix~\ref{app:user_simulator}). A hypothetical pure binary protocol would retain only the selected branch and would generally require additional questions. When dictionary lookup fails because extracted names do not match benchmark annotations, an LLM fallback derives an answer from the original description.

\paragraph{Configuration.}
Our main configuration uses a budget of $30$ questions and an initial prior range of $[0.01\times,100\times]$ around the LLM guess. The loop stops when total entropy falls below $-40$, when no progress is recorded for $5$ rounds, or when the budget is spent; a parameter is resolved when its posterior standard deviation is below $10^{-8}$. Direct questions, solver-derived importance, and optional refinement are enabled. Soft refinement uses a blending weight of $0.3$. Confirmation candidates become available below $0.1$ times the current interval width and receive their main score below $0.05$ times that width. Binary splits include the posterior midpoint and quartiles. One note for anyone running our code: because the ablation below shows no clear benefit from solver-derived importance, the released default disables it and assigns uniform importance instead. Reproducing the main configuration requires re-enabling the solver-derived scores.

\paragraph{Importance scoring and question scoring.}
The composite importance score uses weights $(0.40, 0.30, 0.15, 0.15)$ for objective swing, local gradient sensitivity, shadow-price magnitude, and binding-set variation. The finite-difference perturbation is $\max(0.05|\mu|,\,0.1\sigma,\,10^{-6})$. Questions are ranked by expected entropy reduction times an importance-dependent scaling term. Direct questions get a multiplicative bonus of $1.5$, confirmation questions a factor of $0.8$, and a default importance of $0.5$ applies whenever no score is available.
We picked the five weights $(w_{\mathrm{sw}}, w_{\mathrm{gr}}, w_{\mathrm{sh}}, w_{\mathrm{bd}}, \alpha)$ by hand, with two aims: put more weight on objective-level signals than on local gradients and dual information, and keep the importance term $g(s_i){=}\alpha{+}(1{-}\alpha)s_i \in [0.5,1]$ from swamping entropy reduction. We did not tune them on the benchmarks. This conservative choice may be part of why importance scoring shows such a small average effect in the ablation, and a systematic sweep over these weights and over $\alpha$ would be worth running.

\paragraph{Metrics.}
We report exact objective-value agreement, agreement within $1\%$, $5\%$, and $10\%$, mean objective error, resolved-literal rate, and average question count. ``Exact'' means objective error below $0.01\%$. We cap each instance's error at $100\%$ when computing the mean so that a few failures do not dominate. The resolved-literal rate is $(\textsc{Known}+\textsc{Filled})/\text{all extracted literals}$; it is not the fraction of masked ground-truth parameters recovered. The primary objective comparison uses the benchmark answer. Re-solving the translated model after substituting benchmark literals provides a separate internal diagnostic. Because the $100\%$ cap conceals the error tail, we report tolerance rates alongside the mean.

\subsection{Main Results}

\begin{table*}[t]
\centering
\small
\caption{Objective-value agreement across datasets. The \textbf{No-interact.} row solves from the translator's initial literals without questions. Deltas are percentage-point differences. ``Exact'' denotes error below $0.01\%$; ``Gap'' is mean objective error capped at $100\%$ per instance; ``Resolved'' is $(\textsc{Known}+\textsc{Filled})/\text{all extracted literals}$.}
\label{tab:main_results_transposed}
\renewcommand{\arraystretch}{1.08}
\setlength{\tabcolsep}{3pt}
\begin{tabularx}{\textwidth}{@{}l
>{\raggedright\arraybackslash}X
>{\raggedright\arraybackslash}X
>{\raggedright\arraybackslash}X
>{\raggedright\arraybackslash}X
>{\raggedright\arraybackslash}X
>{\raggedright\arraybackslash}X
>{\raggedright\arraybackslash}X@{}}
\toprule
\textbf{Metric} & \textbf{NL4LP} & \textbf{EasyLP} & \textbf{NL4OPT} & \textbf{ReSocratic} & \textbf{IndustryOR} & \textbf{ComplexOR} & \textbf{ComplexLP} \\
\midrule
\rowcolor{gray!12} \textbf{Exact}
& \textbf{87.6\%} 
& \textbf{79.8\%} 
& \textbf{78.1\%} 
& \textbf{77.9\%} 
& \textbf{64.3\%} 
& \textbf{50.0\%} 
& \textbf{27.0\%}  \\

No-interact.
& 16.3 {\scriptsize\textcolor{red}{$\downarrow$71.3}}
& 34.2 {\scriptsize\textcolor{red}{$\downarrow$45.6}}
& 13.2 {\scriptsize\textcolor{red}{$\downarrow$64.9}}
& 13.1 {\scriptsize\textcolor{red}{$\downarrow$64.8}}
& 15.0 {\scriptsize\textcolor{red}{$\downarrow$49.3}}
& 5.6 {\scriptsize\textcolor{red}{$\downarrow$44.4}}
& 5.7 {\scriptsize\textcolor{red}{$\downarrow$21.3}} \\
\addlinespace[2pt]

$\leq$1\%
& 88.2\% {\scriptsize\textcolor{green!50!black}{$\uparrow$0.6}}
& 80.2\% {\scriptsize\textcolor{green!50!black}{$\uparrow$0.4}}
& 80.7\% {\scriptsize\textcolor{green!50!black}{$\uparrow$2.6}}
& 81.6\% {\scriptsize\textcolor{green!50!black}{$\uparrow$3.7}}
& 64.3\% {\scriptsize\textcolor{green!50!black}{$\uparrow$0.0}}
& 50.0\% {\scriptsize\textcolor{green!50!black}{$\uparrow$0.0}}
& 32.2\% {\scriptsize\textcolor{green!50!black}{$\uparrow$5.2}} \\

$\leq$5\%
& 90.4\% {\scriptsize\textcolor{green!50!black}{$\uparrow$2.8}}
& 82.2\% {\scriptsize\textcolor{green!50!black}{$\uparrow$2.4}}
& 85.1\% {\scriptsize\textcolor{green!50!black}{$\uparrow$7.0}}
& 83.9\% {\scriptsize\textcolor{green!50!black}{$\uparrow$6.0}}
& 64.3\% {\scriptsize\textcolor{green!50!black}{$\uparrow$0.0}}
& 55.6\% {\scriptsize\textcolor{green!50!black}{$\uparrow$5.6}}
& 53.1\% {\scriptsize\textcolor{green!50!black}{$\uparrow$26.1}} \\

$\leq$10\%
& 91.6\% {\scriptsize\textcolor{green!50!black}{$\uparrow$4.0}}
& 83.1\% {\scriptsize\textcolor{green!50!black}{$\uparrow$3.3}}
& 87.3\% {\scriptsize\textcolor{green!50!black}{$\uparrow$9.2}}
& 86.6\% {\scriptsize\textcolor{green!50!black}{$\uparrow$8.7}}
& 66.7\% {\scriptsize\textcolor{green!50!black}{$\uparrow$2.4}}
& 61.1\% {\scriptsize\textcolor{green!50!black}{$\uparrow$11.1}}
& 62.6\% {\scriptsize\textcolor{green!50!black}{$\uparrow$35.6}} \\

Gap
& 5.5\%
& 12.0\%
& 8.1\%
& 8.0\%
& 22.7\%
& 21.1\%
& 25.2\% \\

Resolved
& 96.4\%
& 93.0\%
& 96.0\%
& 97.2\%
& 88.5\%
& 85.6\%
& 88.1\% \\

Avg Q
& 1.4
& 1.5
& 1.4
& 1.8
& 2.4
& 1.9
& 5.7 \\

\bottomrule
\end{tabularx}
\end{table*}

\paragraph{Aggregate results.}
Table~\ref{tab:main_results_transposed} shows higher objective-value agreement under the interactive configuration than under the no-interaction baseline in every dataset, although the comparison also includes optional refinement and therefore does not isolate question selection. Exact agreement ranges from $87.6\%$ on NL4LP to $27.0\%$ on ComplexLP, while average question counts range from $1.4$ to $5.7$. Wider tolerances increase agreement, especially on ComplexLP. These values measure agreement with the benchmark objective under an idealized simulator; they are not literal-recovery or optimizer-recovery rates.

\subsection{Ablation Study}

We examine solver-assisted prioritization, refinement, prior range, question budget, and translation model.

\paragraph{Solver-aware prioritization features.}
Table~\ref{tab:ablation_exact} gives exact objective-value agreement for \sailor\ and two
ablated variants. \emph{W/o importance} turns off solver-derived scoring
and picks questions by entropy reduction alone; \emph{w/o refinement}
turns off the optional LLM refinement step.

Removing importance scoring changes results in both directions across datasets. The unweighted macro-average changes from $66.4\%$ to $66.1\%$, providing no clear macro-level benefit. The small IndustryOR and ComplexOR samples make their dataset-level differences uncertain. The linear weighting $g(s_i){=}\alpha{+}(1{-}\alpha)s_i$ with $\alpha{=}0.5$ also limits the magnitude of reweighting. Since the solver perturbations cost extra solves without buying accuracy here, we set uniform importance as the released default and treat solver-derived scoring as an option worth revisiting once its weights are tuned.

\paragraph{Refinement step.}
Removing refinement lowers the reported rate on four datasets, leaves one unchanged, and raises it on two. The unweighted macro-average changes from $66.4\%$ to $64.2\%$. This mixed pattern is exploratory: it does not identify a general refinement effect without paired records and a factorial design. Refinement remains externally conditioned because it follows a revealed benchmark value rather than free-running self-correction.

\begin{table*}[t]
\centering
\small
\caption{Ablation: exact objective-value agreement
for \sailor\ and two ablated variants. Deltas are percentage-point changes
relative to \sailor; the final column is an unweighted macro-average.}
\label{tab:ablation_exact}
\renewcommand{\arraystretch}{1.12}
\setlength{\tabcolsep}{3pt}
\begin{tabularx}{\textwidth}{@{}l
>{\centering\arraybackslash}X
>{\centering\arraybackslash}X
>{\centering\arraybackslash}X
>{\centering\arraybackslash}X
>{\centering\arraybackslash}X
>{\centering\arraybackslash}X
>{\centering\arraybackslash}X
|>{\centering\arraybackslash}X@{}}
\toprule
\textbf{Configuration}
  & \textbf{NL4LP} & \textbf{EasyLP} & \textbf{NL4OPT}
  & \textbf{ReSoc} & \textbf{IndOR} & \textbf{CmplxOR}
  & \textbf{CmplxLP} & \textbf{Macro avg.} \\
\midrule
\rowcolor{gray!10}
\textbf{\sailor}
  & 87.6 & 79.8 & \textbf{78.1}
  & \textbf{77.9} & \textbf{64.3} & \textbf{50.0}
  & 27.0 & 66.4 \\
w/o importance
  & \textbf{89.9}\,{\scriptsize\textcolor{green!50!black}{$\uparrow$2.3}}
  & \textbf{81.3}\,{\scriptsize\textcolor{green!50!black}{$\uparrow$1.5}}
  & 77.6\,{\scriptsize\textcolor{red}{$\downarrow$0.5}}
  & 76.7\,{\scriptsize\textcolor{red}{$\downarrow$1.2}}
  & 61.9\,{\scriptsize\textcolor{red}{$\downarrow$2.4}}
  & 44.4\,{\scriptsize\textcolor{red}{$\downarrow$5.6}}
  & \textbf{30.8}\,{\scriptsize\textcolor{green!50!black}{$\uparrow$3.8}}
  & 66.1\,{\scriptsize\textcolor{red}{$\downarrow$0.3}} \\
w/o refinement
  & 86.5\,{\scriptsize\textcolor{red}{$\downarrow$1.1}}
  & 80.7\,{\scriptsize\textcolor{green!50!black}{$\uparrow$0.9}}
  & 76.8\,{\scriptsize\textcolor{red}{$\downarrow$1.3}}
  & 77.9\,{\scriptsize\textcolor{gray}{$\approx$0}}
  & 59.5\,{\scriptsize\textcolor{red}{$\downarrow$4.8}}
  & 38.9\,{\scriptsize\textcolor{red}{$\downarrow$11.1}}
  & 29.4\,{\scriptsize\textcolor{green!50!black}{$\uparrow$2.4}}
  & 64.2\,{\scriptsize\textcolor{red}{$\downarrow$2.2}} \\
\bottomrule
\end{tabularx}
\end{table*}

\begin{figure}[htb]
    \centering
    \includegraphics[width=\textwidth]{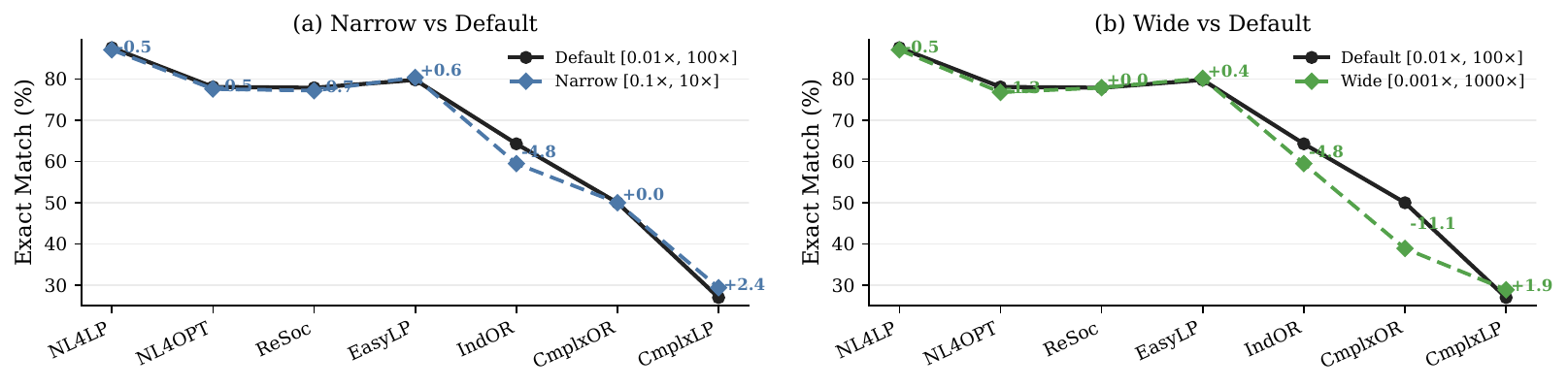}
    \caption{Sensitivity of exact objective-value agreement to the prior range.
    Narrow and wide priors behave inconsistently across datasets; annotations
    give aggregate percentage-point differences from the default.}
    \label{fig:ablation_prior}
\end{figure}

\paragraph{Prior range sensitivity.}
Figure~\ref{fig:ablation_prior} varies the prior range around the
initial guess. The reported differences are sensitivity estimates, not
precise effects: narrow and wide ranges change results in both directions
across datasets. Moreover, multiplicative ranges are not suitable for
negative or zero-valued parameters, as discussed in the method limitations.

\begin{figure}[htb]
    \centering
    \includegraphics[width=\columnwidth]{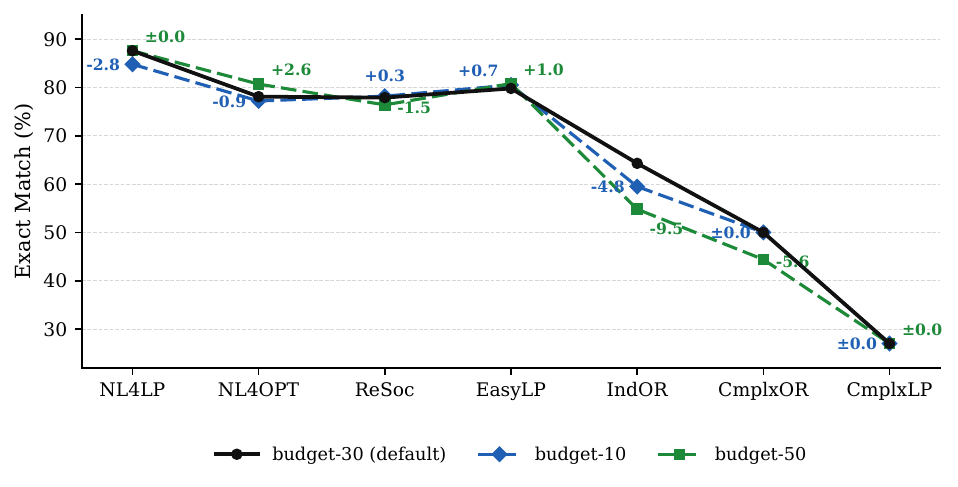}
    \caption{Sensitivity of exact objective-value agreement to question budgets
    $10$ and $50$ relative to the default budget $30$. Effects are mixed
    across datasets.}
    \label{fig:ablation_budget}
\end{figure}

\paragraph{Question budget.}
Figure~\ref{fig:ablation_budget} varies the per-instance question cap
($B \in \{10, 30, 50\}$, default $30$). The differences
are small on several datasets and mixed elsewhere. ComplexLP's agreement
within $10\%$ rises from $58.8\%$ at budget $10$ to $62.6\%$ at budget
$30$, while its reported ``Exact'' agreement remains $27.0\%$. Budget $50$ reports
lower values on IndustryOR and ComplexOR than budget $30$. The mechanism
has not been isolated.

\begin{figure}[htb]
    \centering
    \includegraphics[width=\columnwidth]{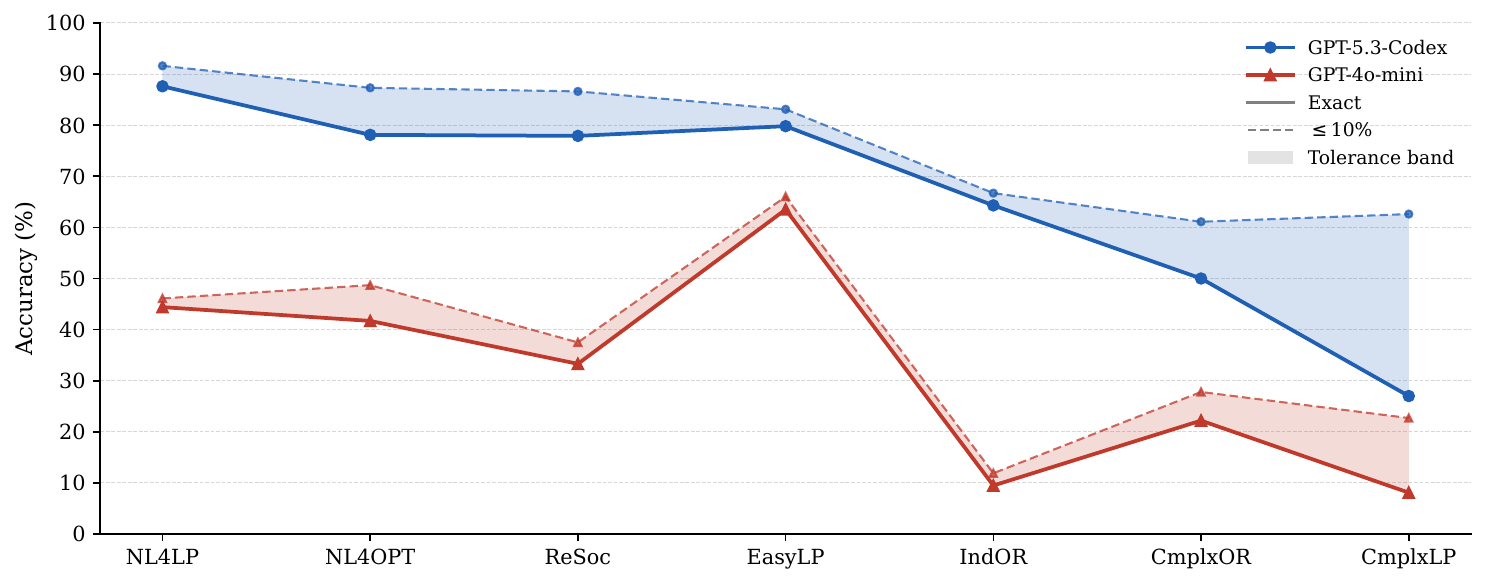}
    \caption{Objective-value agreement for two translation
    models. Solid lines show exact agreement; shaded bands extend
    to agreement within $10\%$.}
    \label{fig:llm_ablation}
\end{figure}

\paragraph{Translation model choice.}
Last, we swap GPT-5.3-Codex for GPT-4o-mini as the translator.
Figure~\ref{fig:llm_ablation} reports the largest differences among the
shown ablations. The reported ``Exact'' agreement falls from $87.6\%$ to $44.4\%$ on
NL4LP, from $64.3\%$ to $9.5\%$ on IndustryOR, and from $50.0\%$ to
$22.2\%$ on ComplexOR. Solver-assisted interaction cannot repair an
optimization structure that was wrong to begin with.

\section{Discussion}

The experiment provides a proof of concept for an optimization assistant that does not require users to anticipate every numerical input. Rather than treating an executable first translation as final, the system identifies unsupported choices and asks about the ones that most affect the current model. Objective-value agreement improves over the initial-guess baseline in every dataset, but the feedback is unusually informative, so the question counts do not transfer directly to human interaction.

The ablations do not establish that solver-derived importance improves aggregate performance. Its unweighted macro-average is slightly lower than the full configuration and dataset effects have mixed signs, which is why our released default falls back to uniform importance. Refinement is also mixed and remains exploratory. Translation-model choice produces larger reported differences, consistent with the structural limitation that literal updates cannot repair an incorrectly translated formulation.

\subsection{Limitations and threats to validity}
\label{sec:limitations}

\paragraph{Idealized interaction.}
The simulator answers deterministically and reveals the exact value after direct, binary, and confirmation questions. Real users may answer approximately, decline, or provide inconsistent units. The study is therefore a controlled branch-and-reveal evaluation, not evidence about deployed behavior or question efficiency with people.

\paragraph{No-interaction baseline.}
The no-question baseline keeps the translator's initial literals. The full configuration adds both interaction and optional refinement, so their effects are not isolated. A paired factorial evaluation is needed.

\paragraph{Metric scope.}
The primary metric compares the recovered objective value with the benchmark answer. It establishes neither literal-by-literal recovery, nor equivalence of feasible regions, nor recovery of a particular optimizer, and an internal re-solve of the translated model answers a different diagnostic question. We evaluate $1{,}723$ of $1{,}732$ source instances after the nine exclusions described above.

\paragraph{Benchmark-construction circularity.}
The masking procedure uses objective sensitivity, solution sensitivity, and feasibility impact, overlapping with the recovery-time importance estimates. This favors literals the method is designed to prioritize. Independent masking by structural role or random selection among extractable literals is needed.

\paragraph{Small datasets.}
IndustryOR and ComplexOR contain only $42$ and $18$ source instances, respectively. Their dataset-level and ablation differences should therefore be treated as descriptive.

\paragraph{Translator dependence and structural mismatch.}
Our results assume a capable translation model; with GPT-4o-mini the pipeline degrades badly, so the approach as reported is tied to a strong and relatively expensive model, and translator strength is entangled with the value of the interactive machinery in a way these experiments do not separate. Interactive updates repair numerical errors but cannot repair an optimization structure the initial translation got wrong. Coupling structural model repair with solver-assisted parameter recovery is an important direction for future work.

\paragraph{Belief-model mismatch.}
Raw-scale Gaussian beliefs and multiplicative intervals do not adequately represent negative, zero, bounded, or integer-valued literals. Type-aware supports and likelihoods are required before applying the method broadly.

\paragraph{Comparison with full input.}
On ComplexLP, objective-value agreement is $46.0\%$ with fully specified input and $27.0\%$ with the interactive masked-input configuration. These regimes are not directly comparable, but the difference cautions against presenting interaction as a general improvement over complete specifications.

\section{Conclusion}

We presented \sailor, a proof-of-concept optimization assistant for users who omit required numbers from an initial natural-language description. The system detects unsupported numerical choices, asks targeted clarification questions, rewrites the translated model, and re-solves. Under an idealized simulator, this process produces higher objective-value agreement than retaining the translator's initial guesses. Our ablations locate the binding constraint in translation rather than in elicitation: changing the translation model moves results by tens of percentage points, whereas removing solver-derived question prioritization produces a difference too small for our sample sizes to resolve. The evaluation does not establish performance with human users, pure binary feedback, or structurally incorrect models. Future work should evaluate whether real users understand and can answer the questions, support approximate or range-valued responses, measure literal, objective, feasibility, and decision outcomes independently, and separate the value of question ordering from the value of the answers it elicits.

\bibliographystyle{tmlr}
\bibliography{ref}
\newpage
\appendix

\section{Masking Implementation Details}
\begin{figure}[htb]
\centering

\begin{exampleblock}[origbg]{Original problem description}
A factory produces two types of food, I and II, and currently has
\textbf{50} skilled workers. One skilled worker can produce
\textbf{10 kg/h} of food I or \textbf{6 kg/h} of food II.
A worker works \textbf{40 h} per week.
The weekly wage of a skilled worker is \textbf{360 yuan}.
Workers assigned to overtime work \textbf{60 h} per week with a weekly wage of
\textbf{540 yuan}. If food cannot be delivered on time, the delay penalty is
\textbf{0.5 yuan/kg} for food I and \textbf{0.6 yuan/kg} for food II.
\end{exampleblock}

\vspace{1.5mm}

\begin{exampleblock}[maskbg]{Masked version}
A factory produces two types of food, I and II, and currently has
{\color{maskred}\textbf{some quantity}} skilled workers.
One skilled worker can produce
{\color{maskred}\textbf{a given rate}} of food I or
{\color{maskred}\textbf{a given rate}} of food II.
A worker works {\color{maskred}\textbf{some value}} per week.
The weekly wage of a skilled worker is
{\color{maskred}\textbf{a certain wage}}.
Workers assigned to overtime work
{\color{maskred}\textbf{some value}} per week with a weekly wage of
{\color{maskred}\textbf{a certain wage}}.
If food cannot be delivered on time, the delay penalty is
\textbf{0.5 yuan/kg} for food I and
{\color{maskred}\textbf{some}} for food II.
\end{exampleblock}

\vspace{1.5mm}

\begin{exampleblock}[rephbg]{Rephrased masked version}
A factory produces two types of food, I and II, and currently has
{\color{rephgreen}\textbf{skilled workers on staff}}.
One skilled worker can produce food I at
{\color{rephgreen}\textbf{a rate measured in kg/h}} or food II at
{\color{rephgreen}\textbf{another rate measured in kg/h}}.
Each worker has
{\color{rephgreen}\textbf{a fixed weekly number of working hours}}.
The weekly wage of a skilled worker is
{\color{rephgreen}\textbf{specified in yuan}}.
Some workers may also be assigned to overtime, with
{\color{rephgreen}\textbf{a longer weekly schedule}} and
{\color{rephgreen}\textbf{a corresponding weekly wage}}.
If food cannot be delivered on time, the delay penalty is
\textbf{0.5 yuan/kg} for food I and
{\color{rephgreen}\textbf{a stated amount}} for food II.
\end{exampleblock}

\caption{Our masking pipeline. We take the original description, hide the decision-critical numbers, and optionally paraphrase the result so the underspecified version reads naturally.}
\label{fig:masking_example}
\end{figure}
This appendix gives implementation details for the masking procedure that turns fully specified problems into underspecified ones. The aim is to produce challenging, meaningful instances by hiding numbers that are both structurally important and consequential for the outcome.

\subsection{Code-Level Parameter Extraction via AST}

Given a fully specified model $\mathcal{P}(\boldsymbol{\theta}^{\ast})$, we first get an executable form, for instance Python code using Gurobi. We then parse it with an abstract syntax tree and pull out every numerical constant.

Each constant carries:
\begin{itemize}
    \item its variable name,
    \item its location in the code,
    \item its role in the model, such as objective coefficient, constraint bound, or penalty,
    \item its structural context, meaning whether it sits in the objective, a constraint, or an auxiliary expression.
\end{itemize}

Structured constants such as lists, arrays, and dictionaries are flattened into individual scalars, which lets us mask at a fine grain.

The result is a candidate parameter set
\[
\boldsymbol{\theta}^{\ast} = (\theta_1^{\ast}, \dots, \theta_n^{\ast}),
\]
plus metadata describing each parameter's structural role.

\subsection{Structural Classification of Parameters}

We sort extracted parameters by their role in the model:
\begin{itemize}
    \item \textbf{Objective coefficients}, appearing in $f(x;\boldsymbol{\theta})$,
    \item \textbf{Constraint bounds}, including right-hand sides and limits defining $\mathcal{X}(\boldsymbol{\theta})$,
    \item \textbf{Capacity and demand parameters}, which drive feasibility in many structured problems,
    \item \textbf{Auxiliary constants}, such as big-M values, scaling factors, or numerical stabilizers.
\end{itemize}

This keeps masking diverse and steers effort away from literals estimated to have little effect on the translated model. Because objective coefficients can change both the objective value and the selected optimizer, their effects must be checked rather than inferred from structural role alone.

\subsection{Solver-Based Importance Estimation for Masking}

To keep the masked instances nontrivial, we estimate each parameter's importance by solver-based perturbation. The analysis perturbs each parameter by the same multiplicative factors $\times\{0.5, 0.8, 1.2, 2.0\}$ used in the main text, then folds objective sensitivity, solution sensitivity, and feasibility impact into a single score. The expressions below spell out the objective-sensitivity part.

For each $\theta_i$, we measure the effect on the optimal objective by perturbing and re-solving:
\begin{equation}
g_i =
\frac{
\left|
z^{\ast}\!\left(\boldsymbol{\theta}^{\ast} + \delta_i e_i\right)
-
z^{\ast}\!\left(\boldsymbol{\theta}^{\ast} - \delta_i e_i\right)
\right|
}{
2\delta_i
},
\end{equation}
where $\delta_i > 0$ is a small perturbation and $e_i$ the $i$th standard basis vector.

We also look at objective variation over a wider range:
\begin{equation}
\Delta_i =
\left|
z^{\ast}\!\left(\boldsymbol{\theta}^{\ast}_{i \leftarrow \ell_i}\right)
-
z^{\ast}\!\left(\boldsymbol{\theta}^{\ast}_{i \leftarrow u_i}\right)
\right|,
\end{equation}
where $[\ell_i, u_i]$ is a plausible range around $\theta_i^{\ast}$.

Finally we check whether $\theta_i$ shows up in active or binding constraints, and whether those constraints carry significant dual values. Parameters touching binding constraints or feasibility get higher importance.

These signals combine into a score $\mathrm{Imp}_i$, used only to guide masking during benchmark construction.

\subsection{Masking Policy}

The masking operator picks which parameters to hide, based on importance scores and structural roles.

We do not mask uniformly at random. In our ratio-based policy, candidates with text spans are ranked by sensitivity score, with qualitative importance as a fallback, and the top fraction is selected:
\begin{itemize}
    \item \textbf{Importance driven selection}: higher $\mathrm{Imp}_i$ means more likely to be masked.
    \item \textbf{Nontriviality}: parameters with negligible impact are left alone.
\end{itemize}

Formally, the operator produces
\begin{equation}
\boldsymbol{\theta}^{\ast} \rightarrow (\boldsymbol{\theta}^{\text{obs}}, \boldsymbol{\theta}^{\text{hidden}}),
\end{equation}
where $\boldsymbol{\theta}^{\text{hidden}}$ holds the parameters removed from the observed description.

\subsection{Generation of Underspecified Instances}

Once the hidden set is chosen, we edit the description to remove or obscure the corresponding values. The result is an underspecified description $I_0$ that keeps the structural template but drops the key numbers.

We pair each instance with its ground truth vector $\boldsymbol{\theta}^{\ast}$, which is what makes controlled evaluation through $\mathrm{ErrObj}(\cdot,\cdot)$ possible.

\subsection{Discussion}

The masking procedure pairs code-level analysis with solver perturbations to avoid masking only inert literals. Section~\ref{sec:limitations} discusses the circularity this shares with the recovery-time importance scorer.

\section{Automated User Simulator}
\label{app:user_simulator}

We evaluate \sailor\ with an automated simulator standing in for the human user. The simulator holds the ground-truth vector $\boldsymbol{\theta}^{\ast}$; the recovery system does not. This supports controlled evaluation of question selection, belief updates, and solver-assisted recovery.

\subsection{Simulator Protocol}

At each step the system picks a question $q$ for an unresolved parameter $\theta_i$. The simulator answers from the ground-truth value $\theta_i^{\ast}$, after which the pipeline updates the belief, rewrites the AST, and re-solves.

The simulator handles three question types. For direct value questions it returns the exact value of $\theta_i^{\ast}$. For binary range questions it reports whether $\theta_i^{\ast}\leq m$ or $\theta_i^{\ast}>m$ and then reveals $\theta_i^{\ast}$; the implementation therefore applies an exact update rather than retaining a truncated branch. For confirmation questions it checks whether the proposed value $v$ falls within
\[
\epsilon = 0.1 \times \max(|v|,1.0)
\]
of the truth. Either way, yes or no, it reveals the exact value and the parameter is resolved.

\subsection{Ground Truth Lookup}

LLM-generated parameter names often differ from the benchmark annotations, so lookup runs in three stages. First, an exact dictionary match. If that fails, fuzzy matching on name similarity, description similarity, token overlap, and numerical consistency with the LLM's initial guess. If neither gives a confident match, an LLM fallback gets the original unmasked description and extracts the value.

If all three fail, the question is marked unanswerable. After three consecutive failures on the same parameter, the system locks it at its current best estimate and moves on. This failure-driven fallback is separate from the solver-importance-based automatic locking, which is off in our main configuration.

\subsection{Fidelity and Limitations}

The simulator provides a controlled exact-feedback reference point. Unlike a real user, it has perfect recall, answers deterministically, and returns structured numbers. Real users give approximate, incomplete, or inconsistent responses. The protocol simplifies answer interpretation but does not isolate all error sources: translation, parameter matching, the LLM fallback, belief modeling, and code rewriting can all affect the outcome. A real user might also volunteer information that the simulator withholds.

\end{document}